\documentclass[a4paper,conference]{IEEEtran}
\usepackage{etoolbox}
\makeatletter
\patchcmd{\@makecaption}
  {\scshape}
  {}
  {}
  {}
\makeatother
\usepackage{caption}
\ifCLASSINFOpdf
\usepackage[pdftex]{graphicx}
\else
\fi
\usepackage{amsmath, amssymb}

\usepackage{graphicx}
\usepackage{amsmath}
\usepackage{amssymb}
\usepackage{float}
\usepackage{bm}
\usepackage[T1]{fontenc}
\usepackage[ruled,vlined]{algorithm2e}
\usepackage{color}

\begin{document}
\title{Particle Filter Bridge Interpolation }
\author{\IEEEauthorblockN{Adam Lindhe}
\IEEEauthorblockA{Department of Mathematics\\
KTH Royal Institute of Technology\\
100 44 Stockholm, SWEDEN\\
Email: adlindhe@kth.se}
\and
\and
\IEEEauthorblockN{Carl Ringqvist}
\IEEEauthorblockA{Department of Mathematics\\
KTH Royal Institute of Technology\\
100 44 Stockholm, SWEDEN\\
Email: carrin@kth.se}
\and
\IEEEauthorblockN{Henrik Hult}
\IEEEauthorblockA{Department of Mathematics\\
KTH Royal Institute of Technology\\
100 44 Stockholm, SWEDEN\\
Email: hult@kth.se}
}

\maketitle

\begin{abstract}
Auto encoding models have been extensively studied in recent years. They provide an efficient framework for sample generation, as well as for analysing feature learning. Furthermore, they are efficient in performing interpolations between data-points in semantically meaningful ways. In this paper, we build further on a previously introduced method for generating canonical, dimension independent, stochastic interpolations. Here, the distribution of interpolation paths is represented as the distribution of a bridge process constructed from an artificial random data generating process in the latent space, having the prior distribution as its invariant distribution. As a result the stochastic interpolation paths tend to reside in regions of the latent space where the prior has high mass. This is a desirable feature since, generally, such areas produce semantically meaningful samples. In this paper, we extend the bridge process method by introducing a discriminator network that accurately identifies areas of high latent representation density. The discriminator network is incorporated as a change of measure of the underlying bridge process and sampling of interpolation paths is implemented using sequential Monte Carlo. The resulting sampling procedure allows for greater variability in interpolation paths and stronger drift towards areas of high data density. 
\end{abstract}

\section{Introduction}
The generation of a semantically meaningful transition from one data point to another is commonly referred to as \textit{Interpolation}. The most common example is arguably the construction of a smooth realistic transition from one image to another. Other examples include imputation of missing data and generation of time-series data from snapshots. If data is generated from a probability distribution $p(x)$ of large dimension, that is concentrated within a low-dimensional manifold, then the interpolation path must reside near that manifold in order to produce semantically meaningful samples.

In order to understand the underlying geometry and characterise good interpolation paths, it is useful to learn low-dimensional representations of data. Auto Encoding models such as the Variational Auto Encoder (VAE) \cite{Kingma2013} are effective tools in this context. The VAE consists of an encoder $q_{\phi}(z|x)$ mapping data to a latent space $Z$, a prior distribution $p(z)$ that define the distribution of latent representations, and a decoder,  $p_{\theta}(x|z)$ reconstructing latent representations to the data space. The assignment of latent data representations allows for easier analysis of the data manifold. More specifically, decoded interpolation paths from the latent space that respects the manifold defined by $p(z)$ should respect the manifold defined by $p(x)$ and hence produce semantically meaningful results. Interpolation between two data-points $x^{(i)}$ and $x^{(j)}$ in the VAE context is performed in the following manner:
\\
\begin{enumerate}
    \label{interpolation_algorithm}
    \item Encode $x^{(i)}, x^{(j)}$ through sampling  $z^{(i)} \sim q_{\phi}(z|x^{(i)}), z^{(j)} \sim q_{\phi}(z|x^{(j)})$ 
    \item Construct a suitable path in the latent space between the decoded points
    $[z^{(i)}, z,..., z^{(j)}]$
    \item Decode the latent path by sampling 
    \begin{align*}
        [x^{(i)} \sim p_{\theta}(x|z^{(i)}), x \sim p_{\theta}(x|z),..., x^{(j)} \sim p_{\theta}(x|z^{(j)})]
    \end{align*}
\end{enumerate}

\hspace{10pt}

To reduce noise, the sampling steps (1) and (3) can be replaced by means. Several different methods for constructing optimal paths in step (2) have been suggested and are discussed in Section \ref{relatedwork}. In this article, we are specifically concerned with constructing random interpolations in this setting. That is, constructing a probability distribution over interpolation paths given a set of endpoints, and methods for sampling from that distribution. Sampling random interpolation paths is useful for understanding the space of realistic interpolations, for generating video samples from snapshots, motion samples from static positions, etc. In this article, an extension of the Gaussian Interpolation method presented in \cite{henrikcarl} is presented. 

To model the distribution over interpolation paths in $Z$, they consider an artificial data generating mechanism, in the form of a stationary stochastic process $\{\mathbf{Z}(t)\}_{t \in [0,T]}$ with $p(z)$ as its invariant distribution. The distribution over interpolation paths is given by the distribution of the corresponding bridge process, obtained by conditioning $\mathbf{Z}$ on the start and end points, $\mathbf{Z}(0) = z^{(i)}$ and $\mathbf{Z}(T) = z^{(j)}$. This approach forces latent data representations within the interpolation path to be placed within the manifold where $p(z)$ is concentrated, hence assuring variability in samples and adherence to data location. However, the manifold specified by $p(z)$ can potentially be large, and only partially populated with latent data representations; as there is no mechanism in the setup that specifies where \textit{within} the manifold that data representations are placed.

In this paper we investigate the introduction of such a mechanism in the interpolation sampling procedure. This is achieved through training a discriminator network on the latent space; that distinguishes between samples randomly drawn from $p(z)$, and original data representations. The discriminator network is then applied for constructing a change of measure that emphasizes paths with proximity to data. The change of measure is implemented using a Sequential Monte Carlo method (SMC) where the mutation of particles follows the underlying bridge process and the resampling weights are determined by the discriminator network.  Although it is possible to apply the method to any stochastic process interpolation scheme, we are focusing in this paper on Gaussian Processes and normally distributed priors for demonstration purposes. We demonstrate theoretically and empirically that the method is capable of adhering to the data manifold more effectively than Gaussian Interpolation. This results in better quality samples, and more variability in interpolation paths. 

The paper is organised as follows. In Section \ref{relatedwork}, related work is presented. In Section \ref{bridge}, the Gaussian Process and the corresponding Gaussian Bridge used for interpolation is introduced.  The main results are given in Section \ref{sampling} where the algorithm for adjusting the stochastic bridge process according to data representation proximity is presented, and its mathematical properties established. Numerical experiments on two data sets are provided in Section \ref{experiments}.

\section{Related work} \label{relatedwork}

Previous work on the construction of semantically meaningful interpolation paths is almost entirely devoted to deterministic methods. A useful approach is to find paths that are optimised with respect to some evaluation criteria, such as geodesics with respect to an appropriate metrics. Such methods are developed within the field of manifold learning where interpolation paths are obtained by minimising a cost functional of the form $\int_0^1 f(\gamma(t), \dot \gamma(t)) dt$ over smooth paths $\gamma$. This approach includes the $f$-geodesics by \cite{AvL12}, the method of directed embeddings of \cite{PJM11, PJM14}, and the hitting-time approach of \cite{HSJ15}. Recent mathematical developments of this approach are provided by \cite{DS19}.

In latent variable models, more precisely VAEs, interpolation paths are constructed between encoded data-points in the latent space and the entire path is decoded in the data space. The decoder is generally only able to produce meaningful output from a vicinity of encoded data-points in the latent space. 

Since VAEs are trained to put encoded data-points in the typical set of the prior it is useful to account for the prior in the construction of interpolation paths. For standard normal priors on a high dimensional latent space, encode data-points tend to lie near a sphere around the origin with radius given by the square-root of the dimension, see  \cite{Hall2005}. To account for the prior \cite{White2016} suggests using spherical interpolation as a replacement to linear interpolation, which empirically gives improvements in sample quality. 

A problem with linear interpolation in VAEs is the distributional mismatch identified by \cite{Kilcher2018}, where an alternative prior distribution is suggested on basis of exhibiting lower distributional mismatch. In a similar direction \cite{Agustsson2019}, suggests a warping technique based on optimal transport to reduce the distributional mismatch, a Cauchy prior is proposed by \cite{Lesniak2019}, which is shown to produce a perfect distributional match, and \cite{Singh2019} use numerical optimisation to search for a prior with low  distributional mismatch. The approach using distributional mismatch clearly demonstrates the problems with interpolations not respecting the prior. However, low distributional mismatch will lead to interpolation paths that are good on average,  which is a rather weak criteria. 


A different approach to interpolation is suggested by \cite{Berthelot2018}, where an interpolation quality critic network is included in the training process. Visual improvements are obtained, but the method does not include analysis of the latent variable manifold. 


Gaussian processes are considered as priors of VAEs for dependent data, see \cite{Casale2018}, where the covariance function models the dependence in the training data. 
A similar approach is used for multivariate time series imputation in \cite{FBRM19}. Extensions of the Gaussian process framework for time series include the use of recurrent network encoders and decoders for representation learning, classification and forecasting, see \cite{SMS15, HZG17,LM18}. Stochastic process VAEs are further developed in the ODE$^2$VAE model by \cite{Y}. These methods aim to model distributions for dependent data, such as time series, but are generally not used to construct interpolation paths.  

The approach taken in \cite{henrikcarl} is different from previous approaches to interpolation, mentioned above, which rely on deterministic methods. Stochastic interpolation is introduced by representing the distribution of interpolation paths in the form of associated stochastic bridge processes in the latent space. Although there are connections both to manifold learning and to VAEs for dependent data; none of these existing approaches are used to study interpolation.

In this paper, we build further on the construction of stochastic interpolation paths based on Gaussian bridges, by considering a discriminator network that identifies the vicinity of encoded data-points in the latent space and a change of measure technique, implemented by sequential Monte Carlo, that guides interpolation paths through the relevant part of the latent space.

\section{Gaussian processes and Gaussian bridges}
\label{bridge}
A stochastic process $\{Z(t)\}$ is a centered stationary Gaussian process if for any finite number of times $(t_0,...,t_m)$, it holds that
\begin{align}
\label{gaussianprocess}
    (Z(t_0),...,Z(t_m)) \sim N(\mathbf 0, \Sigma)
\end{align}
 where $\Sigma$ is a covariance matrix. Typically, $\Sigma$ is constructed through setting $\Sigma_{ij} = k(t_i-t_j)$ for some kernel function $k$ fulfilling $k(t_i-t_j) = k(t_j- t_i)$. In this paper, the kernel

\begin{align}
    \label{nonperiodickernel}
     k(h) =  \exp \Big\{ -\beta |h|^{\alpha} \Big\}
\end{align}
 is used. This is the kernel used in \cite{henrikcarl} for non-periodic sampling. The kernel controls the level of covariance between samples in the process path, as function of their difference in time. Large $\alpha$ results in large covariance between close sample points, while small covariance for distant sample points. Low $\alpha$ results in near constant covariance over time distance. Thus $\alpha$ controls the smoothness of paths, while $\beta$ acts as a scale parameter. Note that the process has the standard normal as stationary distribution, making it a suitable alternative for constructing interpolation bridge processes in the context of VAE with normal prior. The bridge is constructed through using the well established rules for conditional distributions of multivariate normals. Given a normal distribution on form (\ref{gaussianprocess}), with $t_{0}=0, t_{m}=T$, successive sampling from $Z(t_k) | Z(t_{0}),..., Z(t_{k-1}), Z(t_{m})$, $k=1,...,m-1$, amounts to successive sampling from dependent normal distributions, and yields a sample interpolation path. Here $T$ controls the level of path randomness. A small $T$ essentially reproduces linear interpolation of 0 variance, and large $T$ essentially reproduces the non-conditioned distribution (\ref{gaussianprocess}) for the path mid-points \cite{henrikcarl}.

\section{Sampling with discriminator network}
\label{sampling}
As discussed in the introduction, the approach taken in \cite{henrikcarl} can potentially fail, or at least produce sub-optimal results, due to data not being evenly spread out over the manifold defined by $p(z)$. The method only assures that the interpolation path resides within the manifold specified by $p(z)$, but it does not specify closer where within the manifold the process should reside; which is a problem in the case data representations only occupies a small part of the manifold. In order to remedy this limitation, we propose training a discriminator network in identifying areas of high data latent representation density; and subsequently use it for causing a drift of the stochastic process within the manifold towards data intensive areas through a sampling process. Areas with high data latent representation is captured by the probability density $L^n(z)$ defined by
\begin{equation*}
    L^N(z) = \frac{1}{N}\sum_{i=1}^n q_\phi(z|x^{(i)}).
\end{equation*}
The discriminatory network is trained to distinguish between the empirical measure $L^N(z)$ and the prior $p(z)$. For example, fix a function class $\mathcal{F}$ and set the discriminatory network $\hat{f}$ to be the maximizer to the following maximization problem
\begin{equation*}
    \max_{f\in\mathcal{F}} \int_Z f(z)L^N(z)dz - \int_Z f(z)p(z)dz. 
\end{equation*}
If $\mathcal{F}$ is the set of Lipschitz continuous functions then the value of $\int_Z \hat{f}(z)L^N(z)dz - \int_Z \hat{f}(z)p(z)dz$ is the Wasserstein distance between $L^N(z)$ and $p(z)$. Another approach is to let $\mathcal{F}$ be all functions parametrized by a neural network with a sigmoid output and then solve the following maximization problem
\begin{equation*}
    \max_{f\in\mathcal{F}} \int_Z \log(f(z))L^N(z)dz + \int_Z\log(1-f(z))p(z)dz.
\end{equation*}
Call the function maximizing the above optimization problem $\hat{f}$. The object now is to interpolate through regions where $\hat{f}(z)$ has a high value. This is achieved by a change of measure of the Gaussian bridge process, implemented via a sequential Monte Carlo scheme.   

\subsection{A change of measure and sequential Monte Carlo}

Let $\mathbb{P}$ be a probability measure such that, under $\mathbb{P}$, the process $\{\hat Z(t)\}_{t \in [0,T]}$ is a Gaussian bridge process, obtained from a centered stationary Gaussian process, $\{Z(t)\}_{t \in \mathbb{R}}$, with kernel $k$ conditioned on $Z(0) = z_0$ and $Z(T)  = z_T$. Given a function $\hat f$ whose value represents the importance of points in the latent space, a terminal time $T > 0$, and a bounded  function $\gamma: [0,T] \to [0,\infty)$, let $\mathbb{Q}^\gamma$ be the probability measure such that 
\begin{align*}
    \frac{d\mathbb{Q}^\gamma}{d\mathbb{P}}(\hat Z) \propto \exp\left(\int_0^T \gamma(t) \log \hat f(\hat Z(t)) dt \right). 
\end{align*}
Note that, compared to $\mathbb{P}$, the measure $\mathbb{Q}^\gamma$ emphasizes paths with a high value of $\hat f$ and the function $\gamma$ is a parameter that controls how much high values of $\hat f$ are emphasized.   

To sample an interpolation path from $\mathbb{Q}^\gamma$, consider sampling the path at a finite number of times $0 = t_0 \leq t_1 \leq \dots \leq t_m = T$.  Let $\mathcal{G}_t$ be the $\sigma$-field generated by $\hat Z(s)$, $0 \leq s \leq t$, and $\mathbb{Q}^\gamma_t$ denote the conditional measure of $\mathbb{Q}^\gamma$ given $\mathcal{G}_t$, and similar for $\mathbb{P}_t$ . Then, the likelihood ratio $\frac{d\mathbb{Q}_{t_{k-1}}}{d\mathbb{P}_{t_{k-1}}}$ restricted to $\mathcal{G}_{t_k}$ is given by
\begin{align*}
   \frac{e^{\int_{t_{k-1}}^{t_k} \gamma(t) \log f(\hat Z(t))) dt} E^{\mathbb{P}}[e^{\int_{t_k}^T \gamma(t) \log \hat f(\hat Z(t)) dt} \mid \mathcal{G}_{t_{k}}] }{E^{\mathbb{P}}[e^{\int_{t_{k-1}}^T \gamma(t) \log \hat f(\hat Z(t)) dt}]}, 
\end{align*}
for $k=1, \dots, m$. To sample from $\mathbb{Q}^\gamma_{t_{k-1}}(\hat Z_{t_k} \in \cdot)$ it is, in principle, possible to sample a proposed Gaussian bridge $\{\hat Z(s), t_{k-1} \leq s \leq t_k\}$ from $\mathbb{P}_{t_{k-1}}$ and accept the proposal with probability proportional to the weight 
\begin{align*}
    w_k \! = e^{\int_{t_{k-1}}^{t_k} \gamma(t) \log \hat f(\hat Z(t)) dt} E^{\mathbb{P}}[e^{\int_{t_k}^T \gamma(t) \log \hat f(\hat Z(t)) dt} \!\mid \mathcal{G}_{t_{k}}]. 
\end{align*}
In practice, a Sequential Monte Carlo implementation is preferred. Given $N$ trajectories $(\hat Z^n(t_0), \dots, \hat Z^n(t_{k-1}))$, $n = 1, \dots, N$, each trajectory is extended to $(\hat Z^n(t_0), \dots, \hat Z^n(t_{k}))$ by sampling each $\hat Z^n(t_k)$ independently from $P(\hat Z^n(t_k) \in \cdot \mid \hat Z^n(t_0), \dots, \hat Z^n(t_{k-1}))$. The extended trajectories are resampled using weights $w_k^n$, where the integrals are approximated using a convex combination of the function value at its end points,
\begin{align*}
    \int_{t_{k-1}}^{t_k} g(t) dt \approx [\xi g(t_{k-1}) + (1-\xi) g(t_k)](t_k - t_{k-1}),
\end{align*}
some $\xi \in [0,1]$. The resulting approximation of the resampling weights is given by
\begin{align*}
    w_k^n &= e^{\int_{t_{k-1}}^{t_k} \gamma(t) \log \hat f(\hat Z^n(t)) dt}  \\ & \quad \times E^{\mathbb{P}}[e^{\int_{t_k}^T \gamma(t) \log \hat f(\hat Z^n(t)) dt} \mid \mathcal{G}_{t_{k}}] \\ 
    &\approx
     [\hat f(\hat Z^n(t_{k-1}))^{\xi \gamma(t_{k-1})}\hat f(\hat Z^n(t_{k}))^{(1-\xi)\gamma(t_{k})}]^{(t_k-t_{k-1})} \\ & \quad  \times 
      [\hat f(\hat Z^n(t_{k}))^{\xi \gamma(t_{k})}\hat f(\hat Z^n(T))^{(1-\xi)\gamma(T)}]^{(T-t_{k})}. 
\end{align*}
In Algorithm \ref{samplingprocess} the sequential Monte Carlo implementation is presented, with equidistant times, $t_k - t_{k-1} = \Delta$, $\xi = 0$ and $\gamma \equiv 1/\Delta$, leading to $w_k^n \propto \hat f(\hat Z(t_k))$. Note that, in Algorithm \ref{samplingprocess}, the number of columns in the variable "paths" is the number of sample paths created, whereas the number of rows is $T/\Delta+1$.

\begin{algorithm}
{\bfseries Input:} time $T$ \\
{\bfseries Input:} number of steps $steps$ \\
{\bfseries Input:} kernel covariance matrix $\Sigma$ \\
{\bfseries Input:} start latent data representation $z_{0}$ \\
{\bfseries Input:} end latent data representation $z_{T}$ \\
\hspace{0pt}
\\
times = linspace(0, T, steps) \\
\hspace{0pt}
\\
paths = $\begin{bmatrix}
z_{0} & z_{0} & \hdots & z_{0}\\
\vdots & \vdots &  & \vdots\\
z_{T} & z_{T} & \hdots & z_{T}\\
\end{bmatrix}$
\hspace{0pt}
\\
\hspace{0pt}
\\
\hspace{0pt}
\\
for i in (1 : times - 1): \\
\quad next\_steps = [] \\
\quad for j in (0 : paths.no\_columns): \\
\quad \quad next\_step $\sim
N(z_{i}^{j}|z_{0:i-1}^{j}, z_{T}, \Sigma)$ \\
\quad \quad next\_steps.append(next\_step)\\
\quad paths[i, :] = next\_steps \\
\quad values = discriminator(next\_steps) \\
\quad weights = values / sum(values) \\
\quad resample\_index $\sim$ Multinomial(\\
\quad \quad N=paths.no\_columns,\\
\quad \quad probabilities=weights\\
\quad ) \\
\quad paths = paths[:, resample\_index] \\
interpolation\_path\_example = paths[:, 0] \\
\caption{Stochastic interpolation with discriminator network sampling}
\label{samplingprocess}
\end{algorithm}

\section{Experiments}
\label{experiments}
\subsection*{Lines data set}
The lines dataset was introduced in \cite{Berthelot2018} and consists of $32\times 32 $ pixel grey-scale images with a single 16 pixel line extending from the center. In \cite{Berthelot2018} two interpolation metrics are also introduced, mean-distance and smoothness. For an interpolation path, mean-distance measure the average cosine-distance to the data set along the interpolation path. Smoothness measures the maximum angle difference between two adjacent lines in the interpolation path. The motivation is that the interpolation should go through data and therefore have low cosine distance to the data set, i.e., low mean score,  and the interpolation should be continuous, i.e., no big difference in adjacent interpolation points, which corresponds to low smoothness score. The VAE setup follows the setup of \cite{Berthelot2018}. The encoder consists of 5 blocks where each block consists of two $3\times3$ convolutional layers followed by a $2\times2$ average pooling layer.  The decoder uses five blocks, each block consisting of two $3\times3$ convolutional layers followed by a $2\times 2$ nearest neighbour up-sampling. The latent space dimension is set to 2 to allow for easy visualization of the latent space. 

\begin{figure}[!t]
 \centering
\includegraphics[width=200pt]{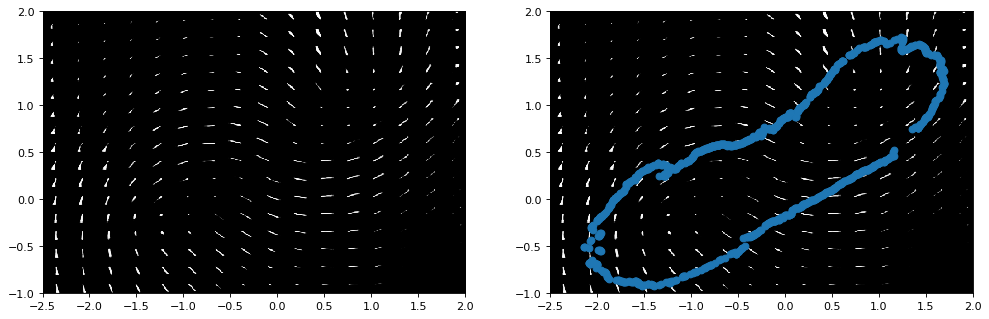}
\captionsetup{font=footnotesize, width= 70mm}
\caption{\label{latent_ellips} Visualization of the latent space and the encoded means of the data. The image to the left visualize the latent space by decoding points in the latent space and showing the corresponding line for each latent point. The image to the right adds the mean of the encoded images, notice that the data is concentrated on a 1-dimensional curve.}
\end{figure}

The latent space is visualized by plotting the decoded images for each point in the latent space, see Figure \ref{latent_ellips}. It is clear from the figure that the latent space has regions were it has bad reconstruction capabilities. To visualize where the data is located in the latent space, 1000 images are transformed via the encoder to the latent space and their mean vector is plotted. The result, right image in Figure \ref{latent_ellips}, shows that the data is located on a 1-dimensional curve. Also to be noted is that the images decoded from regions close to the 1-dimensional curve have reconstructed images of high quality. 

\begin{figure}[!t]
 \centering
\includegraphics[width=200pt]{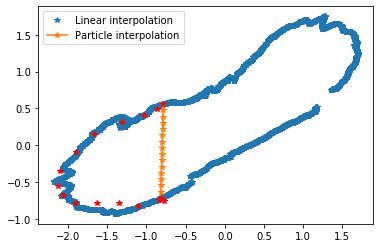}  
\includegraphics[width=200pt]{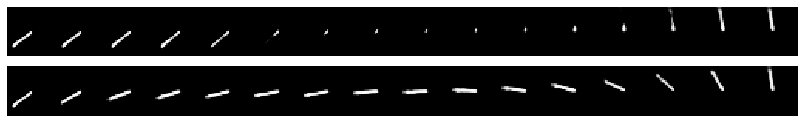}
\captionsetup{font=footnotesize, width= 70mm}
\caption{\label{linear_bad} Top a linear interpolation and a SMC interpolation visualized in the latent space, notice how the linear interpolation passes through a region where no data is located. Middle: The linear interpolation in the image space. Bottom: The SMC interpolation in the image space}
\end{figure}

The structure of the data in the latent space implies that simple linear interpolation often will pass through regions with bad reconstruction. As an example, see Figure \ref{linear_bad}, where the linear interpolation from one side of the curve to the other passes through the center where the reconstruction is poor. The SMC interpolation on the other hand follows the data and creates a smooth interpolation between the two lines. 

For quantitative evaluation of the mean and smoothness score we run 50 interpolations with 16 images in each interpolation. The Gaussian process in the Gaussian process interpolation and for the proposal distribution in the SMC interpolation use the kernel defined in \eqref{nonperiodickernel} with parameters  $\alpha = 2, \beta = 5$. The discriminatory network is a four layer neural network with a sigmoid output. The hidden layers have sizes 100, 200, 500 and use a ReLU activation. The Adam optimizer with binary cross entropy as loss is used for training. 1000 particles are then used for the SMC interpolation.

\begin{figure}[!t]
 \centering
\includegraphics[width=200pt]{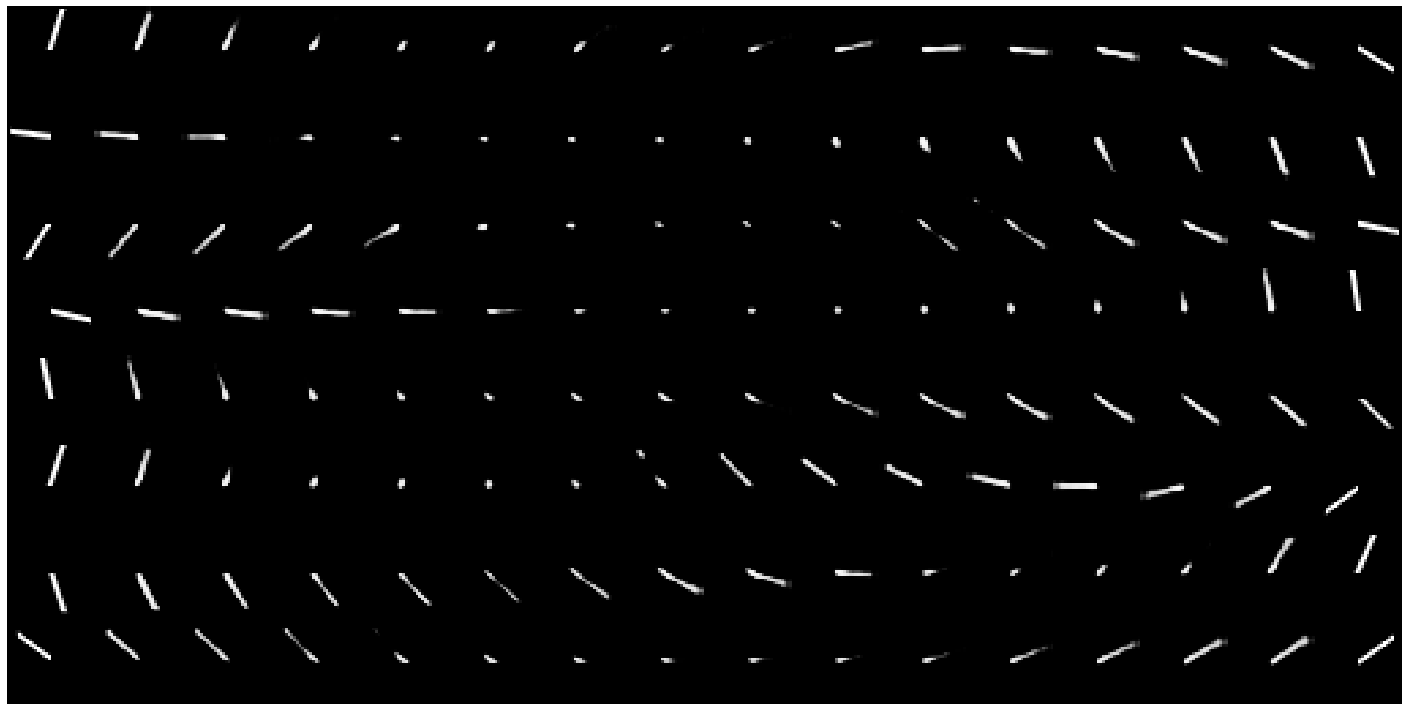}  
\includegraphics[width=200pt]{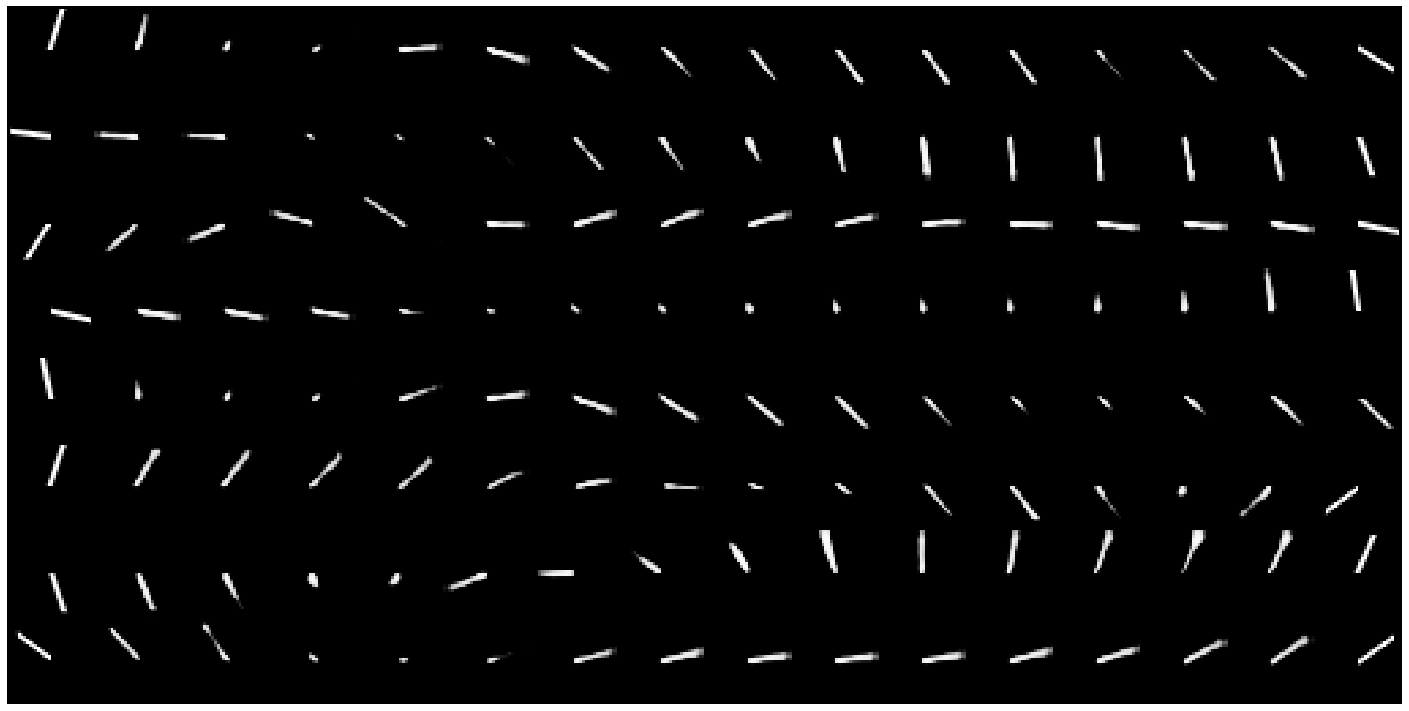}
\includegraphics[width=200pt]{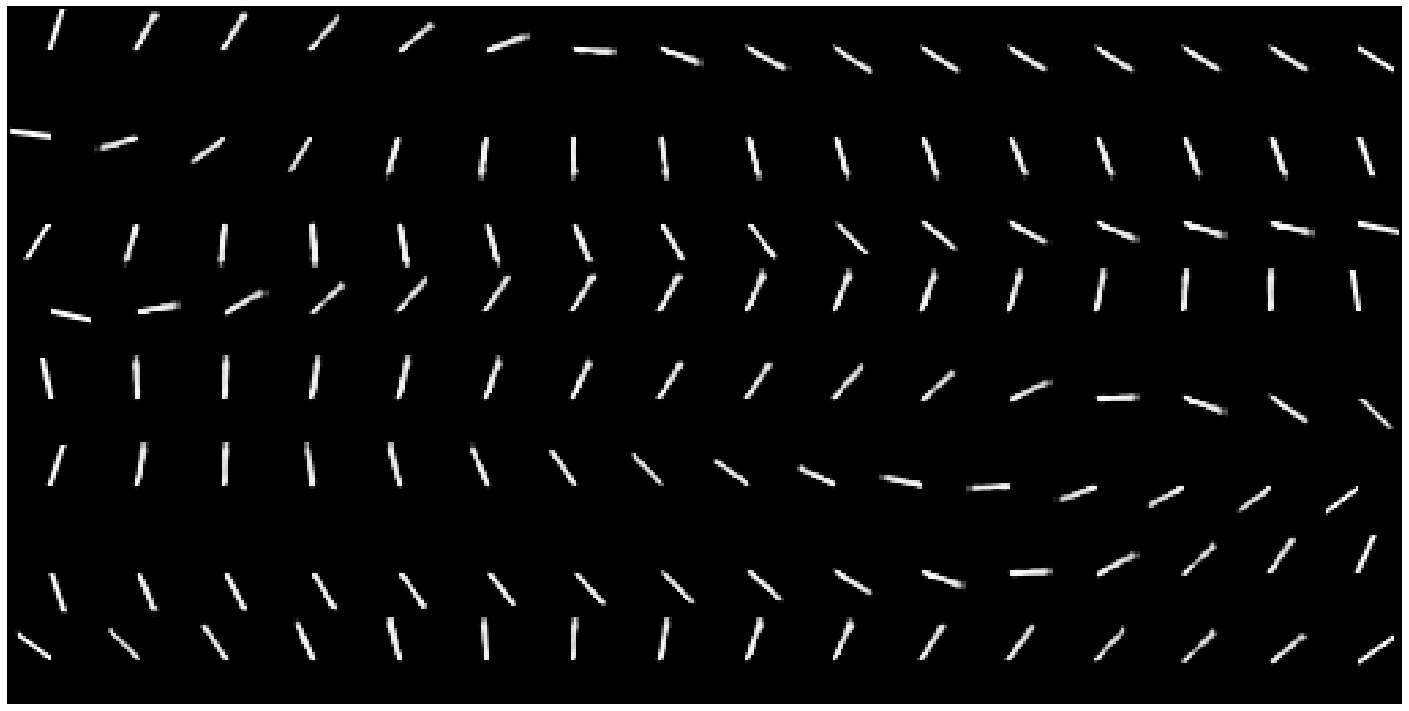}
\captionsetup{font=footnotesize, width= 70mm}
\caption{\label{lines_ex} Interpolation examples in the lines dataset. Top: linear interpolation, middle: Gaussian interpolation, bottom: SMC interpolation.}
\end{figure}

In Table \ref{lines_tab} the mean score and smoothness score for the three different methods, linear, Gaussian and SMC are presented. The SMC method has the lowest score for both the mean score and the smoothness score. Further examples of interpolation with the three different methods are presented in Figure \ref{lines_ex}, the SMC method has the sharpest images and the smoothest paths.    

To investigate performance on a larger latent space, the same experiment is set up on a latent space of dimension 64. The encoder and decoder have similar structure as above but with four blocks instead of five. This is exactly the same setup as in \cite{Berthelot2018}. In Table \ref{lines_tab64} the results from the experiments with latent space dimension of 64 are shown. The results are overall slightly better than for two dimensional latent space and the SMC method perform significantly better at both mean score and smoothness score. 
\begin{table}[h]
    \centering
    \captionsetup{font=footnotesize, width= 70mm}
    \caption{\label{lines_tab} Mean and Smoothness score for Linear, Gaussian and smc interpolation. $\alpha=2$, $\beta=5$, $T=1$, latent space dimension = 2}
    \begin{tabular}{l|c|c}
      Method & Mean score (std) & Smoothness score (std) \\
    \hline
        Linear & 0.0122 (0.0119)& 0.692 (0.480)\\
        Gaussian & 0.0106 (0.0109) & 0.544 (0.411)\\
        SMC & 0.00223 (0.00174) & 0.159 (0.260)\
    \end{tabular}
    
\end{table}

\begin{table}[h]
    \centering
    \captionsetup{font=footnotesize, width= 70mm}
     \caption{\label{lines_tab64}Mean score and Smoothness score for Linear, Gaussian and SMC interpolation. $\alpha=2$, $\beta=5$, $T=1$, latent space dimension = 64}
    \begin{tabular}{l|c|c}
      Method & Mean score (std) & Smoothness score (std) \\
    \hline
        Linear & 0.0105 (0.0113)& 0.346 (0.285)\\
        Gaussian & 0.00868 (0.0101) & 0.411 (0.425)\\
        SMC & 0.00200 (0.00172) & 0.0842 (0.0364)\
    \end{tabular}

\end{table}


\subsection*{CelebA data set}
In order to test the algorithm on more complex data requiring higher prior dimension, we consider the CelebA data set \cite{celeba}; a large-scale face attributes data set with more than 200K celebrity images.

In order to construct sample images of high fidelity, a variant of the VAE model is used, as suggested by X. Hou et al \cite{vggvae}. In this variant, image representations in the first layers of the VGG-net \cite{vggnet} are used as components in the training loss function. The setup used in this article, including network architecture, batch size, loss function, training scheme etc.,  is identical to \cite{vggvae}. The Gaussian Process with kernel (\ref{nonperiodickernel}), with parameters $\alpha$=2, $\beta$=2.5,  is used for both Gaussian Interpolation and SMC Interpolation. These parameters, as well as the chosen times described below, are chosen based on experimental visual inspection of high image quality.

The discriminator network consists of six densely connected layers. The first five layers have ReLU activation functions, and are of unit sizes 500, 300, 100, 50 and 10 respectively. The last layer outputs a single number through a sigmoid function. The Adam optimiser scheme with binary cross entropy loss is used for training, with a batch size of 20000 and 10000 training steps. 
The discriminator network architecture is developed by inspecting how the network output varies between latent space input from "empty" areas and input from data intensive areas. A flexible network results in spikes around latent data points, rendering the sampling mechanism in the SMC Interpolation unable to identify suitable paths towards data as soon as a proposal is not in the immediate vicinity of a latent data representation point. A rigid network results in over regularisation, indicating data is present in empty areas. Smoothly decreasing output when moving from data latent points towards empty areas is preferred; and obtained by the proposed architecture for test samples.

\begin{figure}[!t]
 \centering
\includegraphics[width=200pt]{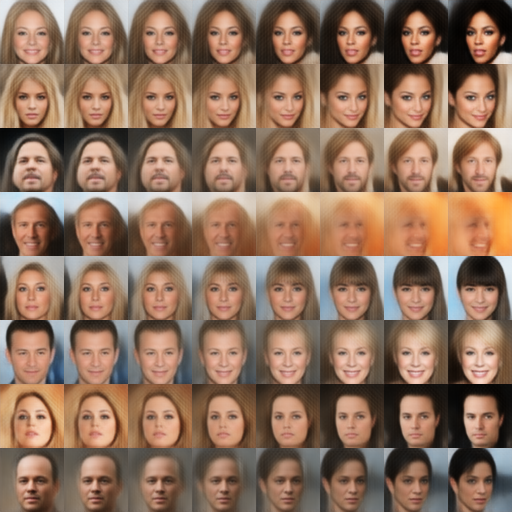}
\captionsetup{font=footnotesize, width= 70mm}
\caption{\label{linearinterpolation} Linear interpolation}
\end{figure}

For reference, eight samples of linear interpolation are shown in Figure \ref{linearinterpolation}. The pictures in the far left and far right columns are reconstructions of data images.

\begin{figure}[]
 \centering
\includegraphics[width=200pt]{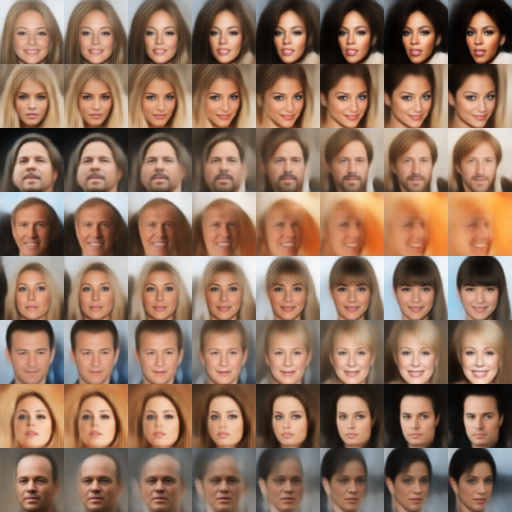}
\captionsetup{font=footnotesize, width= 70mm}
\caption{\label{gaussian_interpolation_t_0_7} Gaussian interpolation, time 0.7}

\vspace{7pt}

\includegraphics[width=200pt]{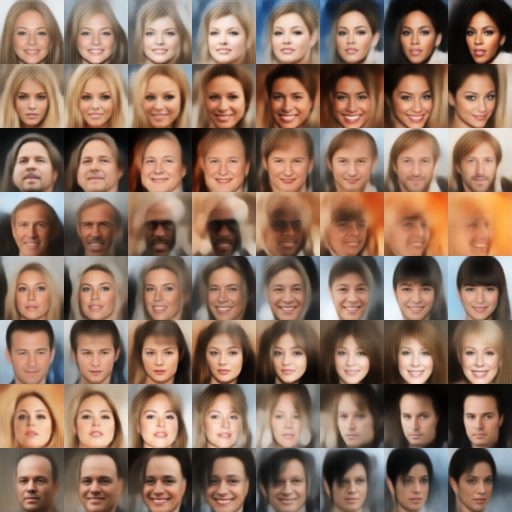}
\captionsetup{font=footnotesize, width= 70mm}
\caption{\label{gaussian_interpolation_t_2} Gaussian interpolation, time 2}

\vspace{7pt}

\includegraphics[width=200pt]{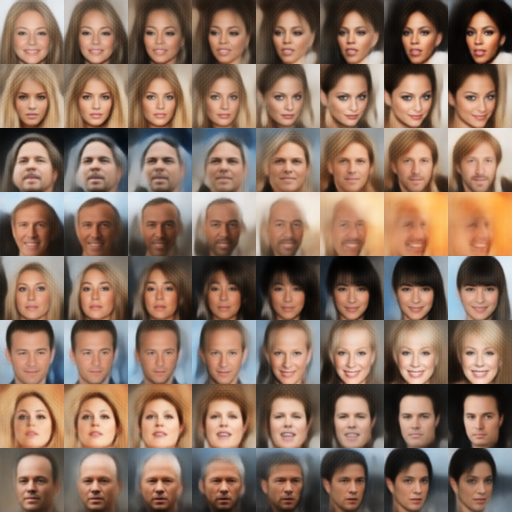}
\captionsetup{font=footnotesize, width= 70mm}
\caption{\label{particle_interpolation_t_2} SMC interpolation, time 2}

\end{figure}

\begin{figure}[H]
 \centering
 
 \includegraphics[width=200pt]{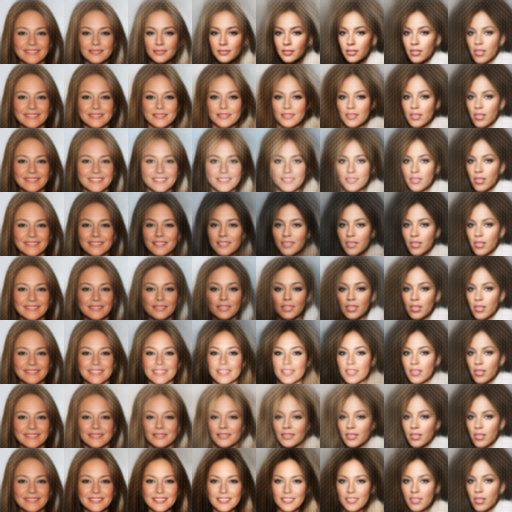}
 \captionsetup{font=footnotesize, width= 70mm}
\caption{\label{gaussian_interpolation_variation_t_0_7} Gaussian interpolation variation, time 0.7}

\vspace{7pt}

\includegraphics[width=200pt]{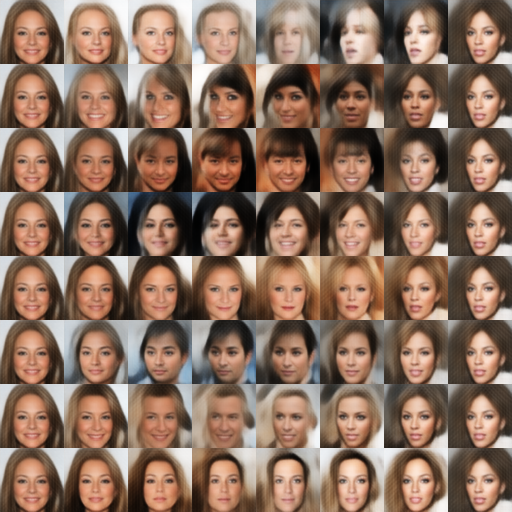}
\captionsetup{font=footnotesize, width= 70mm}
\caption{\label{gaussian_interpolation_variation_t_2} Gaussian interpolation variation, time 2}

\vspace{7pt}

\includegraphics[width=200pt]{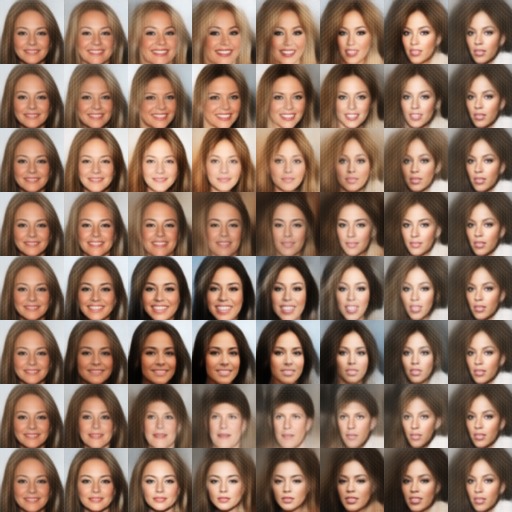}
\captionsetup{font=footnotesize, width= 70mm}
\caption{\label{particle_interpolation_variation_t_2} SMC interpolation variation, time 2}
\end{figure}

Optimizing Gaussian Interpolation time $T$ for visual quality yields a time of around 0.7. The result can be seen in Figure \ref{gaussian_interpolation_t_0_7}. Note the high similarity to the linear samples Fig. \ref{linearinterpolation}. In Figure \ref{gaussian_interpolation_variation_t_0_7}, eight interpolations between the start and end picture of the the first row of Figure \ref{gaussian_interpolation_t_0_7} is displayed. Note that variability between the interpolations is rather limited.

In order to increase variability of samples, $T$ can be increased \cite{henrikcarl}. Increasing the time to $2$ results in higher variability, but degraded image quality, as seen in Figure \ref{gaussian_interpolation_t_2} and Figure \ref{gaussian_interpolation_variation_t_2}.

Extensive inspection of SMC Interpolation for different times and number of particles show no significant visual improvement over Gaussian Interpolation of short time when it comes to image fidelity. However, SMC Interpolation allows for greater variability while maintaining quality. The interpolation results for SMC Interpolation of time 2 is seen in Figure \ref{particle_interpolation_t_2} and Figure \ref{particle_interpolation_variation_t_2}. Here, 200 particles are used. Note that the image quality is again similar to that of Gaussian interpolation of short time, although the variability remains high. Thus our method seems to enable the use of higher time in order to achieve greater variability, with no or limited effect on image quality.

Similar conclusions can be drawn from quantitative metrics. As proxy metric for interpolation image quality, we use the Euclidean distance between the latent representation of the interpolation midpoint and the set of latent representations of data (Mean Score). Note that this mean score is inspired by, but not equivalent to, the mean score suggested in \cite{Berthelot2018}. Averaging the result for a large sample of random interpolations yields a proxy measure of image quality for an interpolation method. As proxy measure for variability of an interpolation method, we use the standard deviation mean over interpolation midpoint coordinates in the latent space. The result using 4000 random interpolations as sample is displayed in Table \ref{quanttable}.

\begin{table}[h]
    \centering
    \captionsetup{font=footnotesize, width= 70mm}
    \caption{\label{quanttable}$\alpha=2$, $\beta=2.5$}
    \begin{tabular}{l|c|c|c}
      Time & Method & Mean Score & Variability Score \\
    \hline
        0.7 & Gaussian & 39.19 & 0.033\\
        0.7 & SMC & 26.47 & 0.013\\
        2 & Gaussian & 89.96 & 0.083\\
        2 & SMC & 41.32 & 0.065
    \end{tabular}
    
\end{table}

\section{Conclusions}
In this paper a method for constructing semantically meaningful stochastic interpolations between data points is suggested and evaluated. The method is based on latent representation of data, and consists of a VAE, a discriminatory network that identifies areas with high concentration of encoded data, and a SMC method to sample interpolation paths. The proposed methodology is useful for (1) generating  interpolation paths of high quality and (2) generating a variety of different interpolation paths at random. The method presented in this paper has the potential to improve interpolations in both situations, compared to deterministic methods in the first case, and to Gaussian Interpolation in the second.

The lines data set demonstrates the first situation. Here, the SMC method results in interpolation paths that adheres perfectly to the data manifold. This is not achieved by neither linear nor Gaussian Interpolation, and it is difficult to imagine any other method, for the specific VAE in question, that could produce better interpolation paths. The CelebA data set demonstrates the second situation. Here, we are interested in producing variety of interpolation paths of high quality. The results demonstrate that the SMC method can produce much higher variability of paths in comparison to Gaussian Interpolation, with limited or no reduction in sample quality. The SMC method comes at an additional computational cost, compared to Gaussian Interpolation, which is proportional to the number of particles used in the SMC algorithm. For applications, we believe that the increase of quality and variability significantly outweighs the cost of additional computation. 

\clearpage

\clearpage

\begin{thebibliography}{99.}%

\bibitem{Agustsson2019}
E.\ Agustsson, A.\ Sage, R.\ Timofte, L.V\ Gool
\newblock Optimal transport maps for distribution preserving operations on latent spaces of generative models.
\newblock  \emph{International Conference on Learning Representations}, 2019.

\bibitem{AvL12}
M.\ Alamgir and U.\ von Luxburg.
\newblock Shortest path distance in random k-nearest neighbor
graphs. 
\newblock \emph{in Proc. 29th Int. Conf. on Machine Learning}, 2012. 


\bibitem{BS14}
C.\ Bayer and J.\ Schoenmakers, J.
\newblock Simulation of forward-reverse stochastic representations
for conditional diffusions. 
\newblock \emph{Ann. Appl. Probab.} 24(5), 1994-2032, 2014. 

\bibitem{Berthelot2018}
D.\ Berthelot, C.\ Raffel, A.\ Roy, and I.\ Goodfellow.
\newblock Understanding and improving interpolation in autoencoders via adversarial regularizer.
\newblock  \emph{Preprint}, arXiv:1807.07543, 2018.

\bibitem{BRSV08}
A.\ Beskos, G.\ Roberts, A.\ Stuart and J.\ Voss, J. 
\newblock MCMC methods for diffusion bridges.
\newblock \emph{Stoch. Dyn.} 8(3), 319–350, 2008.


\bibitem{Casale2018}
F.P.\ Casale, A.V.\ Dalca, L.\ Saglietti, J.\ Listgarten, N.\ Fusi  
\newblock Gaussian Process Variational Autoencoders
\newblock  \emph{Conference and Workshop on Neural Information Processing Systems}, 2018.

\bibitem{Cetin}
U.\ Cetin and A.\ Danilova.
\newblock Markov Bridges: SDE Representation
\newblock  \emph{Stochastic Processes and their Applications}, 126(3):651-679, 2016.


\bibitem{C90}
J.\ Clark, J. 
\newblock The simulation of pinned diffusions. 
\newblock \emph{In Decision and Control, 1990., Proceedings
of the 29th IEEE Conference on}, 1418–1420. IEEE, 1990. 


\bibitem{DS19}
E.\ Davis and S.\ Sethuraman.
\newblock Approximating geodesics via random points.
\newblock \emph{Ann. Appl. Probab.,} 29(3), 1446-1486, 2019. 

\bibitem{DH06}
B.\ Delyon and Y.\ Hu. 
\newblock Simulation of conditioned diffusion and application to parameter
estimation. 
\newblock \emph{Stochast.  Process. Appl.} 116(11), 1660-1675, 2006. 

\bibitem{DG02}
G.B.\ Durham and A.R.\ Gallant.
\newblock Numerical techniques for maximum likelihood estimation
of continuous-time diffusion processes. 
\newblock \emph{J. Bus. Econom. Statist.} 20(3), 297–338, 2002. 

\bibitem{DSV07}
D.\ Gasbarra, T.\ Sottinen and E.\ Valkeila. 
\newblock Gaussian bridges. 
\newblock \emph{In Stochastic analysis and
applications, volume 2 of Abel Symp.}, 361-382, Springer, Berlin, 2007. 

\bibitem{FBRM19}
V.\ Fortuin, D.\ Baranchuck, G.\ R\"atsch, and S.\ Mandt.
\newblock Deep amortized variational inference for multivariate time series imputation with latent Gaussian process models. 
\newblock \emph{2nd Symposium on Advances in Approximate Bayesian Inference}, 1–15, 2019. 

\bibitem{Hall2005}
P.\ Hall, J.S.\ Marron, and A.\ Neeman. 
\newblock Geometric representation of high dimension, low sample size data.
\newblock  \emph{Journal of the Royal Statistical Society: Series B (Stastistical Methodology)}, 67(3):427-444, 2005.

\bibitem{HSJ15}
T.B.\ Hashimoto, Y.\ Sun, and T.S.\ Jaakkola.
\newblock Metric recovery from directed un-weighted graphs. \newblock \emph{in Proc. 18th Int. Conference on AI and Stat (AISTATS)}, San Diego, CA., 2015

\bibitem{fidscore}
M.\ Heusel, H.\ Ramsauer, T.\ Unterthiner, B.\ Nessler, S.\ Hochreiter
\newblock GANs trained by a two time-scale update rule converge to a local nash equilibrium.
\newblock  \emph{NIPS p.p 6629–6640}, 2017.

\bibitem{vggvae}
X.\ Hou, L.\ Shen, K.\ Sun, G.\ Qiu
\newblock Deep Feature Consistent Variational Autoencoder.
\newblock  \emph{IEEE Winter Conference on Applications of Computer Vision (WACV), pp. 1133-1141}, 2017.

\bibitem{HZG17}
W.-N. Hsu, Y. Zhang, and J. Glass. 
\newblock Unsupervised learning of disentangled and interpretable
representations from sequential data. 
\newblock \emph{In Advances in neural information processing systems}, 1878–1889, 2017.





\bibitem{Ionescu2014}
C.\ Ionescu, D.\ Papava, V.\ Olaru, and C.\ Sminchisescu. 
\newblock Human3.6M: Large scale datasets and predictive methods for 3D human sensing in natural environments.
\newblock  \emph{IEEE Transactions on Pattern Analysis and Machine Intelligence}, 36(7):1312-1324, 2014.

\bibitem{KS91}
I.\ Karatzas and S.E.\ Shreve.
\newblock \emph{Brownian motion and stochastic calculus.} 
\newblock  2nd Ed., Springer, 1991.

\bibitem{Kilcher2018}
Y.\ Kilcher, A.\ Lucchi, and T.\ Hofmann.
\newblock Semantic interpolation in implicit models.
\newblock  \emph{International Conference on Learning Representations}, 2018.

\bibitem{Kingma2013}
D.\ Kingma and M.\ Welling.
\newblock Auto-encoding variational Bayes.
\newblock  \emph{Preprint}, arXiv:1312.6114, 2013.

\bibitem{LeCun1998}
Y.\ LeCun, L.\ Bottou, Y.\ Bengio, and P. Haffner.
\newblock Gradient-based learning applied  to  document recognition.
\newblock \emph{Proceedings of the IEEE}, 86(11):2278-2324, 1998.

\bibitem{Lesniak2019}
D.\ Lesniak, I.\ Sieradzki, and I.\ Podolak.
\newblock Distribution-interpolation trade off in generative models.
\newblock  \emph{International Conference on Learning Representations}, 2019.

\bibitem{LM18}
Y.\  Li and S.\ Mandt. 
\newblock Disentangled sequential autoencoder. 
\newblock \emph{ICML}, 2018. 

\bibitem{celeba}
Z.\  Liu, P.\ Luo, X.\ Wang and X.\ Tang. 
\newblock Deep Learning Face Attributes in the Wild. 
\newblock \emph{Proceedings of International Conference on Computer Vision (ICCV)}, 2015. 


\bibitem{PJM11}
D.\ Perrault-Joncas and M.\ Meil\u{a}. 
\newblock Directed Graph Embedding: an Algorithm based on Continuous Limits of Laplacian-type Operators
\newblock \emph{NIPS}, 2011. 

\bibitem{PJM14}
D.\ Perrault-Joncas and M.\ Meil\u{a}. 
\newblock Estimating vector fields on manifolds and the embedding of directed graphs. 
\newblock \emph{arXiv:1406.0013v1}, 2014. 

\bibitem{henrikcarl}
C.\ Ringqvist, H.\ Hult, J.\ Butepage and H.\ Kjellström. 
\newblock Interpolation in Auto Encoders with Bridge Processes.
\newblock \emph{ICPR}, 2020. 

\bibitem{SvdMvZ15}
M.\ Schauer, F.\ van der Meulen and H.\ van Zanten. 
\newblock Guided proposals for simulating multi-dimensional diffusion bridges.
\newblock \emph{Bernoulli}, 23(4A), 2917-2950, 2015. 

\bibitem{vggnet}
K.\ Simonyan and A.\ Zisserman
\newblock Very deep convolutional
networks for large-scale image recognition.
\newblock  \emph{arXiv preprint
arXiv:1409.1556}, 2014.

\bibitem{Singh2019}
R.\ Singh, P.\ Turaga, S.\ Jayasuriya, R.\ Garg, and M.W.\ Braun.
\newblock Non-parametric priors for generative adversarial networks.
\newblock  \emph{International Conference on Machine Learning}, 2019.


\bibitem{SMS15}
N.\  Srivastava, E.\  Mansimov, and R.\ Salakhutdinov. 
\newblock Unsupervised learning of video
representations using lstms. 
\newblock \emph{Proceedings of the 32nd International Conference on International Conference on Machine Learning}, 37, 843–852, 2015. 


\bibitem{Sznitman}
A.S.\ Sznitman.
\newblock Topics in propagation of chaos.
\newblock  \emph{Ecole d'Eté de Probabilités de Saint-Flour XIX. Lecture Notes in Mathematics, vol 1464. Springer, Berlin, Heidelberg} 1989.

\bibitem{TBL18}
M.\ Tschannen, O.\  Bachem, and M.\ Lucic. 
\newblock Recent advances in autoencoder-based representation learning. 
\newblock \emph{arXiv preprint arXiv:1812.05069}, 2018.


\bibitem{White2016}
T.\ White.
\newblock Sampling generative network.
\newblock  \emph{Preprint}, arXiv:1609.04468, 2016.

\bibitem{Y}
C.\ Ylidiz, M.\ Heinonen, and H.\ L\"ahdesm\"aki.
\newblock ODE$^2$VAE: Deep generative second order ODEs with Bayesian neural networks.
\newblock 33rd Conference on Neural Information Processing Systems (NeurIPS), 2019. 

\end{thebibliography}
\end{document}